%% file: eccv2022submission.tex
\documentclass[runningheads]{llncs}
\usepackage{graphicx}
\usepackage{tikz}
\usepackage{comment}
\usepackage{amsmath,amssymb} % define this before the line numbering.
\usepackage{color}

\usepackage{amsmath}
\usepackage{amssymb}
\usepackage{booktabs}
\usepackage{verbatim}
\usepackage{multirow}
\usepackage{enumitem}
\usepackage{dashrule}
\usepackage{subcaption,booktabs}
\usepackage[pagebackref,breaklinks,colorlinks]{hyperref}

\usepackage[accsupp]{axessibility}  % Improves PDF readability for those with disabilities.

\usepackage[width=122mm,left=12mm,paperwidth=146mm,height=193mm,top=12mm,paperheight=217mm]{geometry}

\begin{document}
% \renewcommand\thelinenumber{\color[rgb]{0.2,0.5,0.8}\normalfont\sffamily\scriptsize\arabic{linenumber}\color[rgb]{0,0,0}}
% \renewcommand\makeLineNumber {\hss\thelinenumber\ \hspace{6mm} \rlap{\hskip\textwidth\ \hspace{6.5mm}\thelinenumber}}
% \linenumbers
\pagestyle{headings}
\mainmatter
\def\ECCVSubNumber{615}  % Insert your submission number here

\title{Explicit Occlusion Reasoning for Multi-person 3D Human Pose Estimation} % Replace with your title

% CAMERA READY SUBMISSION
% \begin{comment}
\titlerunning{Explicit Occlusion Reasoning for 3D HPE}
% If the paper title is too long for the running head, you can set
% an abbreviated paper title here
%
\author{Qihao Liu\inst{1} \and
Yi Zhang\inst{1} \and
Song Bai\inst{2\dag} \and
Alan Yuille\inst{1}}
\authorrunning{Q. Liu et al.}
% First names are abbreviated in the running head.
% If there are more than two authors, 'et al.' is used.
%
\institute{Johns Hopkins University \\ \and
ByteDance}
% \institute{Princeton University, Princeton NJ 08544, USA \and
% Springer Heidelberg, Tiergartenstr. 17, 69121 Heidelberg, Germany
% \email{lncs@springer.com}\\
% \url{http://www.springer.com/gp/computer-science/lncs} \and
% ABC Institute, Rupert-Karls-University Heidelberg, Heidelberg, Germany\\
% \email{\{abc,lncs\}@uni-heidelberg.de}}
% \end{comment}
%******************
\maketitle
\def\thefootnote{\dag}\footnotetext{Corresponding author}\def\thefootnote{\arabic{footnote}}
\begin{abstract}

Occlusion poses a great threat to monocular multi-person 3D human pose estimation due to large variability in terms of the shape, appearance, and position of occluders. While existing methods try to handle occlusion with pose priors/constraints, data augmentation, or implicit reasoning, they still fail to generalize to unseen poses or occlusion cases and may make large mistakes when multiple people are present. Inspired by the remarkable ability of humans to infer occluded joints from visible cues, we develop a method to explicitly model this process that significantly improves bottom-up multi-person human pose estimation with or without occlusions. First, we split the task into two subtasks: visible keypoints detection and occluded keypoints reasoning, and propose a Deeply Supervised Encoder Distillation (DSED) network to solve the second one. To train our model, we propose a Skeleton-guided human Shape Fitting (SSF) approach to generate pseudo occlusion labels on the existing datasets, enabling explicit occlusion reasoning. Experiments show that explicitly learning from occlusions improves human pose estimation. In addition, exploiting feature-level information of visible joints allows us to reason about occluded joints more accurately. Our method outperforms both the state-of-the-art top-down and bottom-up methods on several benchmarks. Code is coming soon.
   
\keywords{Human pose estimation, 3D from a single image}
\end{abstract}

\section{Introduction}
\label{sec:intro}
\input{01-intro_v2.tex}

\section{Related Work}
\label{sec:relatedW}
\input{02-related_work_v2.tex}

\section{HUPOR}
\label{sec:occR}
\input{03-occReasoning.tex}

\section{SSF for Occlusion Label Generation}
\label{sec:occL}
\input{04-occLabelGen.tex}

\section{Experiments}
\label{sec:exp}
\input{05-exp_v1.tex}

\section{Conclusions}

Although occlusion is a well-known problem in HPE, not enough attention has been paid to \textit{learning from occlusion}. In this work, we show its value by incorporating our proposed 3D occlusion reasoning method in an existing framework, and present HUPOR. It solves the keypoint detection in a detect-and-reason pipeline, which is effective but being ignored. For occlusion reasoning, we propose a Deeply Supervised Encoder Distillation (DSED) network to effectively infer occluded joints from visible cues, and greatly improve current SOTA 3D and 2D methods. We also propose a Skeleton-guided human Shape Fitting (SSF) method for better human mesh reconstruction and occlusion label generation. Experiments show the effectiveness of our reasoning method in both 2D and 3D HPS. Meanwhile, we demonstrate that both the SSF and the DSED are crucial. 

~\\
\noindent{\textbf{Acknowledgements }}
This work was supported by NIH R01 EY029700. We thank the anonymous reviewers for their efforts and valuable feedback to improve our work.

\clearpage
% ---- Bibliography ----
%
% BibTeX users should specify bibliography style 'splncs04'.
% References will then be sorted and formatted in the correct style.
%
\bibliographystyle{splncs04}
\bibliography{egbib}
\end{document}

%% file: 01-intro_v2.tex
\begin{figure}
    \centering
    \includegraphics[width=0.7\columnwidth]{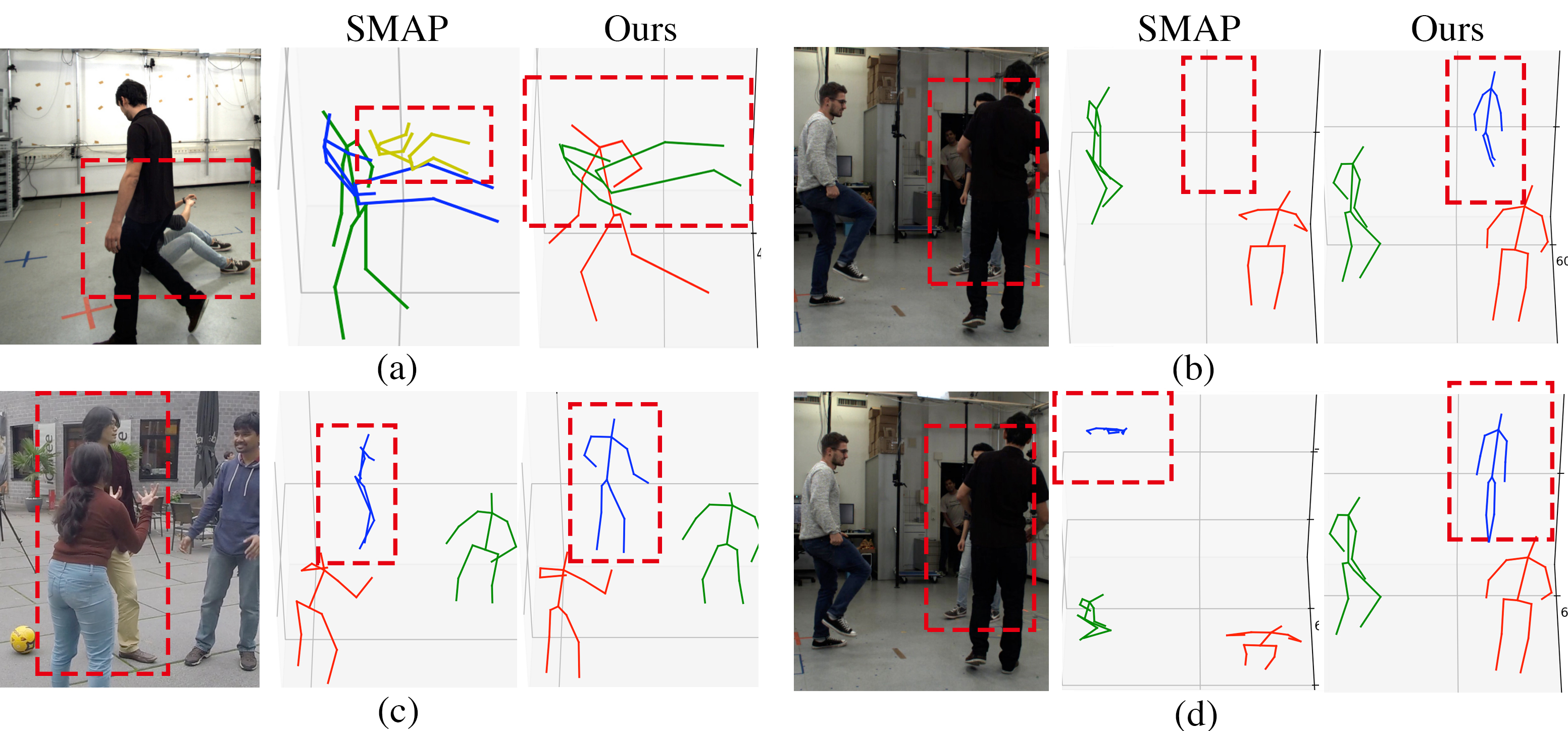}  
    \caption{Current methods may still fail in presence of heavy occlusion, and lead to (a) extra person, (b) missing person, (c) incomplete skeleton, and (d) wrong position estimate. We improve the robustness by enabling the network to infer occluded joints from visible cues explicitly. Red box is only used for visualization.}
    \label{fig:main}
\end{figure}

Monocular 3D multi-person human pose estimation (HPE) is a fundamental task in computer vision with wide applications in robotics, human activation recognition, human-computer interaction,~\etc. Despite great advances brought by neural networks, it remains a very challenging task due to the depth ambiguity, high degrees of freedom in human poses, and frequent occlusion of various forms. Among these challenges, we focus on occlusion which accounts for huge errors in state-of-the-art (SOTA) methods, \eg SMAP~\cite{zhen2020smap}, as shown in Fig.~\ref{fig:main}.

Some existing methods handle occlusion by imposing pose prior and kinematic constraints either explicitly~\cite{radwan2013monocular} or implicitly~\cite{rogez2017lcr}. These priors are learned from a limited training set thus may not generalize well. Others reason about occlusion implicitly through redundant pose encoding~\cite{mehta2018single,mehta2019xnect} or attention mechanism~\cite{kocabas2021pare,gu2021exploring}, where models need to identify occluded joints implicitly without any supervision during training and deal with unreliable intermediate representations with complex conflicts, making learning increasingly difficult.
More importantly, most of these methods, even those designed for multi-person tasks, still solve the problem of occlusion from the perspective of single-person, thus requiring assembling pose first. However, the task of grouping human joints from incomplete keypoints detection is already error-prone. Thus these methods may make large mistakes when multiple people are present. 

% ———————————————————————————— 

To alleviate the occlusion problem in multi-person scenarios, we revisit human's ability to reason about occluded joints. Given an image with occluded people (\eg Fig.~\ref{fig:main}), humans can first precisely `detect' visible joints and visible skeleton information. Then we can reasonably infer the occluded keypoints based on the detected information/cues, combined with local and global contexts, and our prior knowledge on human pose, shape, and motion. We argue that if we allow the network to fully and properly exploit the detected and restructured cues at the feature level, our method should also be able to reason about occluded joints more accurately, just like humans, without bells and whistles. 

To achieve this, we follow the discussion above and split the commonly used single-step keypoint detection into two steps: visible keypoint detection and occluded keypoint reasoning. For the first part, any detection method that provides heatmaps of detected keypoints and skeleton information should work. We directly adopt the SOTA bottom-up method as our detection module. Then based on the intermediate results, we propose a reasoning module that efficiently learns structure information to explicitly infer occluded joints from visible ones. After that, we group both detected and inferred joints into individuals and refine the results. We name this model HUPOR (HUman Pose estimation via Occlusion Reasoning). We show that even using the same detection network and grouping policy, our method significantly improves the ability of 3D bottom-up methods to precisely predict visible keypoints and reasonably infer occluded joints. Furthermore, our method also benefits 2D human pose estimation, by using 2D keypoint detection methods as the detection module and slightly modifying the reasoning module of our method.

% ———————————————————————————— 

Although explicit occlusion reasoning is to some degree intuitive, not enough attention has been paid to it due to the lack of annotations and the poor performance of existing networks (\eg hourglass~\cite{newell2016stacked,li2019rethinking}) in occlusion reasoning. We propose two methods to solve these problems separately:

% ----------------------------

First, explicit occlusion reasoning requires per-joint visibility label which is not available on most 3D human pose datasets. 
To this end, we first fit parametric shape models~\cite{SMPL:2015} to the image and then use projection relationships to determine whether the joint is visible. For human shape fitting, the SOTA 2D/3D pose-based methods~\cite{choi2020pose2mesh} usually fail to predict accurate shape while the image-based methods~\cite{kolotouros2019learning,joo2020exemplar} may get better shape but worse pose estimate and are very sensitive to noise. Therefore, we propose a Skeleton-guided human Shape Fitting (SSF) method that combines the advantage of these two lines of work. From the reconstructed mesh, pseudo occlusion labels are then generated. Unlike previous work~\cite{cheng2019occlusion} which only captures self-occlusion, our method finds self-occlusion, object occlusion, and occlusion by other people. We demonstrate that SSF is crucial to generate accurate occlusion labels for explicit occlusion reasoning, while previous methods are either inaccurate~\cite{cheng2019occlusion} or require manual labeling~\cite{zhang2020object}.

% ———————————————————————————— 
Another challenge is how to effectively reason about occlusions. 
We find that directly training an hourglass model to predict occluded joints from visible joints or images does not yield good results. This is due to the complexity that the network should identify useful information among all input features for reconstructing occluded joints. To solve this problem, we propose a stacked Deeply Supervised Encoder Distillation (DSED) network.
DSED splits the reasoning task into two: reconstruct and imitate, with the help of two encoders. The first encoder acts as a teacher to select the most useful information for occluded joint reconstruction, and the second one learns to extract the same information from just visible cues. 
Compared with the vanilla hourglass model, DSED has a much stronger capability to reason about occluded joints.
Compared with using pre-defined constraints, our method improves the performance by giving the network freedom to extract the information it finds the most useful, leading to better cues for occlusion reasoning. 
More importantly, occlusion reasoning in feature level before assembling pose makes our method more suitable for handling occlusion in multi-person scenarios than previous methods.
% ———————————————————————————— 

To evaluate the performance of our method, we perform experiments on the MuPoTS-3D~\cite{mehta2018single}, MuPoTS-synthOcc, 3DPW~\cite{von2018recovering}, 3DPW-OCC~\cite{von2018recovering,zhang2020object}, CMU Panoptic~\cite{joo2015panoptic}, and Human3.6M~\cite{ionescu2013human3} datasets. The results show that our method yields consistently higher accuracy than the SOTA methods for both occlusion and non-occlusion cases. Our bottom-up method surpasses the SOTA bottom-up methods by 6.0 PCK and SOTA top-down methods by 2.8 PCK on MuPoTS. 
In addition, we apply our DSED-based reasoning module to two recent 2D HPE methods (\ie PifPaf~\cite{kreiss2019pifpaf} and HigherHRNet~\cite{cheng2020higherhrnet}) and we also observe consistent improvements, ranging from 0.9 to 1.7 AP, on COCO~\cite{lin2014microsoft} and CrowdPose~\cite{li2019crowdpose}.

In summary, our contributions are three folds:
\begin{itemize}[topsep=0pt,itemsep=0pt,parsep=0pt]

\item We advance the bottom-up pose estimation paradigm by exploiting its intermediate results for explicit occlusion reasoning, to which not enough attention has been paid before. We demonstrate its efficacy and potential in both 3D and 2D human pose estimation (HPE). 

\item We propose \textbf{DSED} network to infer occluded joints efficiently. It solves the reasoning tasks that hourglass fails. We demonstrate that it is superior in scalability and performance, and is crucial for explicit occlusion reasoning. 

\item Our 3D HPE method, denoted as \textbf{HUPOR}, enables more accurate visible keypoints detection and occluded keypoints reasoning, and significantly outperforms SOTA methods on several benchmarks for both occlusion and non-occlusion cases, while generalizing well to in-the-wild images.

\end{itemize}

%% file: 02-related_work_v2.tex
\noindent\textbf{2D Multi-person Human Pose Estimation. } 
Current methods typically follow one of two paradigms: %bottom-up and top-down. 

\begin{sloppypar}
\textbf{Top-down} methods~\cite{newell2016stacked,fang2017rmpe,he2017mask,chu2017multi,huang2017coarse,papandreou2017towards,chen2018cascaded,xiao2018simple,sun2019deep,moon2019posefix,su2019multi,li2019crowdpose,wang2020graph,zhang2020distribution,huang2020devil,cai2020learning,qiu2020peeking,khirodkar2021multi} split the task into two subtasks: (1) detecting bounding box for all people in the image, and (2) performing single person pose estimator for each detected region. These methods typically work better than bottom-up methods and are currently the best performers on datasets such as COCO~\cite{lin2014microsoft} due to the single human assumption and the pose prior the assumption implies. However, these methods struggle in cases of occlusion~\cite{li2019crowdpose} and interactions~\cite{khirodkar2021multi}.
In addition, these two-step methods lack efficiency due to the need for separate human detectors and the repeat of pose estimation for each bounding box. 
\end{sloppypar}

\textbf{Bottom-up} methods~\cite{newell2016associative,pishchulin2016deepcut,insafutdinov2016deepercut,cao2017realtime,papandreou2018personlab,kreiss2019pifpaf,cheng2020higherhrnet,jin2020differentiable,li2020simple,zhou2019objects,braso2021center} start by first detecting identity-free keypoints over the entire image and then grouping joints into individuals. They are usually superior in speed but inferior in accuracy. To handle the grouping problems, recent work predicts offset fields~\cite{pishchulin2016deepcut,papandreou2018personlab} or part affinity fields~\cite{cao2017realtime,kreiss2019pifpaf}, or uses associative embeddings~\cite{newell2016associative} to get joint relationships. 
Different from the work that only uses PAFs to associate body parts, we also use them for occlusion reasoning since they encode useful context information.

\noindent\textbf{3D Multi-person Human Pose Estimation. } 
In recent years a lot of work focuses on single-person 3D poses~\cite{akhter2015pose,chen20173d,pavlakos2017coarse,mehta2017monocular,sun2017compositional,martinez2017simple,moreno20173d,zhou2017towards,mehta2017vnect,kanazawa2018end,yang20183d,pavllo20193d}. Only a few methods explore multi-person pose estimation:

\textbf{Top-down} methods~\cite{rogez2017lcr,rogez2019lcr,moon2019camera,lin2020hdnet,wang2020hmor} require off-the-shelf human detection, then anchor poses-based classification~\cite{rogez2017lcr,rogez2019lcr} or root-joint localization~\cite{moon2019camera} is performed for better estimation. Recent work exploits the prior knowledge of human size~\cite{lin2020hdnet} or ordinal relations among multiple people~\cite{wang2020hmor} to get more accurate depth estimation. 
However, these methods only focus on depth relations and discards many other useful cues. In addition, similar to the 2D top-down methods, they struggle in cases of occlusion and interactions and lack efficiency.

Although \textbf{bottom-up} methods~\cite{zanfir2018deep,mehta2018single,mehta2019xnect,zhen2020smap} are inferior in accuracy, they have their inherent advantage in handling occlusion using joint relationships.
Mehta~\etal~\cite{mehta2018single} propose occlusion-robust pose-map (ORPM) for full-body pose inference. It utilizes joint location redundancy to infer occluded joints, but can only be applied to extremity joints, and needs a predefined skeleton and an extra read-out process. Recently, Xnect~\cite{mehta2019xnect} encodes joint's immediate local context in the kinematic tree to handle occlusion. However, this method can only use joint locations but fails to use link orientation and other cues. Zhen~\etal~\cite{zhen2020smap} propose a depth-aware part association algorithm to add robustness to occlusion. However, it is only designed for associating body parts but cannot infer occluded joints. In addition, all these methods handle occlusion from the perspective of single-person and require grouping joints into individuals first, which leads to error-prone estimates in multi-person.

\noindent\textbf{Occluded Pose. } 
In addition to the bottom-up methods above~\cite{mehta2018single,zanfir2018deep,mehta2019xnect,mehta2017vnect}, many methods have made good progress in occluded pose estimation. One common way to infer occlusion is to first reconstruct a full-body skeleton and then complete missing joints according to statistical and geometric models~\cite{radwan2013monocular,rogez2017lcr,de2018deep,mehta2019xnect}. However, these methods only work for a single person but as mentioned, are less effective in multi-person scenarios, %only focus on structure information but discard the cue from the image, 
and depend on a library of known poses and structure that is easily biased to training data.
Attention mechanism is introduced to enforce the model to focus on non-occluded areas and thus adding robustness to occlusion~\cite{gu2021exploring,kocabas2021pare,zhou2020occlusion}.
Temporal information~\cite{xu2020deep,veges2020temporal,cheng2019occlusion,wang20203d,parger2021unoc,liu2021graph,artacho2020unipose,cai2019exploiting} is another commonly used cue, but such methods require videos input. 
Using data augmentation~\cite{park2021localizing,xu2021monocular,cheng20203d,peng2018jointly} can also alleviate the problem but with minimal effects compared with other methods and does not fully capture the complexity of occlusions in real world~\cite{kocabas2021pare}.
Some recent methods regress multiple plausible poses~\cite{jahangiri2017generating,li2019generating,biggs20203d,wehrbein2021probabilistic} to handle heavy occlusion, in which almost no cue exists. 
For many occluded cases, human can infer occluded joints from pose prior or local and global context cues. Unlike previous work, we explore explicit modeling of occlusion reasoning.
Recently, Zhang~\etal~\cite{zhang2020object} represent human meshes by UV maps and handle occluded parts reconstruction as image-inpainting problem. The requirement of accurate saliency maps limits the performance and generalization ability, and it is designed for single person scenarios.

%% file: 03-occReasoning.tex
\begin{figure*}[t]
    \centering
    \includegraphics[width=0.9\textwidth]{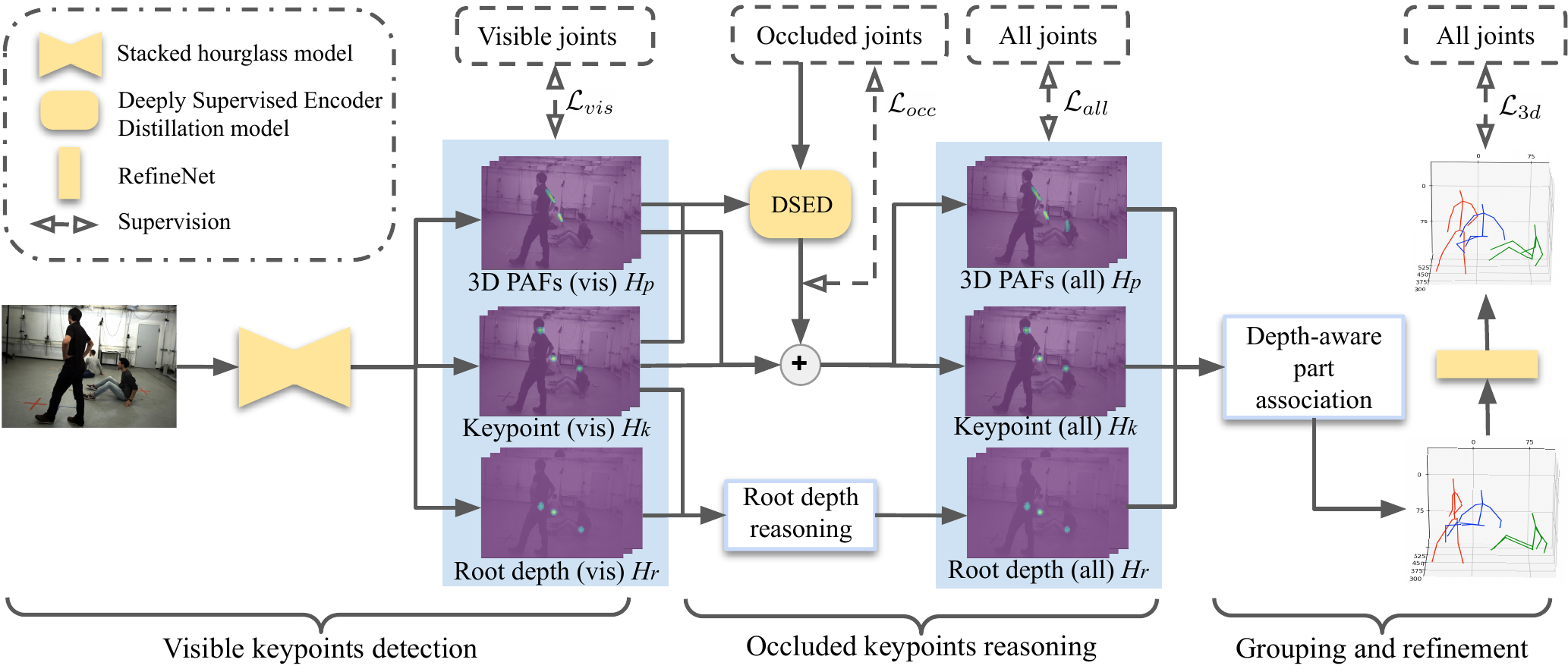}  
    \caption{\textbf{Schematic view}: HUPOR splits the two-step bottom-up methods into three separate tasks: visible keypoint detection, occluded keypoint reasoning, and grouping. In the detection module, a stacked hourglass model~\cite{li2019rethinking,newell2016stacked} is used to extract the intermediate results of all visible joints from the input image. In the reasoning module, DESD network is proposed to better infer occlusion joints from detected visible cues. Finally, all intermediate results are fed to the grouping module to reconstruct the human pose.}
    \label{fig:pipeline}
\end{figure*}
The overall framework of our HUman Pose estimaion network via Occlusion Reasoning (HUPOR) is depicted in Fig.~\ref{fig:pipeline}. It consists of three main stages: given an image $\mathcal{I}\in\mathbb{R}^{4h\times 4w}$, we first use a \textbf{visible keypoints detection} module to detect visible cues, then an \textbf{occluded keypoints reasoning} module is used to reasoning about the invisible information from the visible ones, and finally, all representations are combined to reconstruct 3D poses by the \textbf{grouping and refinement module}. Implementation details are provided in Sup. Mat.

\subsection{Visible Keypoints Detection}
Following SMAP~\cite{zhen2020smap}, we use a stacked hourglass~\cite{li2019rethinking} model to regress several intermediate representations, including 2D keypoint heatmaps, 3D part affinity fields (PAFs), and root depth maps. Let $J$ be the number of joints being considered. Keypoint heatmaps $H_{k}\in\mathbb{R}^{h\times w\times J}$ indicate the probability location of each type of joint for all people in the image. 3D PAFs $H_{p}\in\mathbb{R}^{h\times w\times 3(J-1)}$ are extension of 2D PAFs~\cite{cao2017realtime}. 2D PAFs describe a set of 2D unit vectors pointing from the father node to the child node of the skeleton. For 3D tasks, the relative depth is added to the third dimension. Notice that the first two dimensions are defined in pixel and normalized while the relative depth is in mm without normalization. Root depth maps $H_{r}\in\mathbb{R}^{h\times w\times 7}$ represent the absolute root depth of all people in the images. Different from previous work~\cite{moon2019camera,zhen2020smap} that directly estimates the pelvis depth, we estimate the depth of all the 7 torso joints, including shoulders, pelvis, neck, head, and hips. By doing so, it provides redundant information to infer pelvis depth under occlusion.

During training, we treat the occluded joints as noises and only provide supervision of visible joints. It makes our model more accurate when detecting visible keypoints and PAFs. This module is trained by minimizing $\mathcal{L}_{vis} = \lambda_k^{vis}\cdot\mathcal{L}_{k}^{vis}+\lambda_p^{vis}\cdot\mathcal{L}_{p}^{vis}+\lambda_r^{vis}\cdot\mathcal{L}_{r}^{vis}$, where
\begin{align}
    \mathcal{L}_{k}^{vis} = ||H_{k}-\hat{H}^{vis}_{k}||_2^2 \qquad \mathcal{L}_{p}^{vis} = ||H_p-\hat{H}^{vis}_p||_2^2 \label{eq1}\\
    \mathcal{L}_{r}^{vis} = \sum_{t=1}^7\sum_{i=1}^N||H_{r,t}(u_i^t,v_i^t)-\hat{Z}_i^t||_1
\end{align}
where $\hat{H}^{vis}_{k}$ and $\hat{H}^{vis}_p$ denote ground-truth heatmaps of visible keypoints and visible PAFs, respectively. $(u_i^t,v_i^t)$ is the detected position of the $t^{\text{th}}$ torso joint of the $i^{\text{th}}$ person, $H_{r,t}$ denotes the predicted depth map of torso joint $t$ (\ie, root depth map), and $\hat{Z}$ is the ground-truth normalized depth.

%% ===============================================
\subsection{Occluded Keypoints Reasoning}

\begin{figure}[t]
    \centering
    \includegraphics[width=0.9\columnwidth]{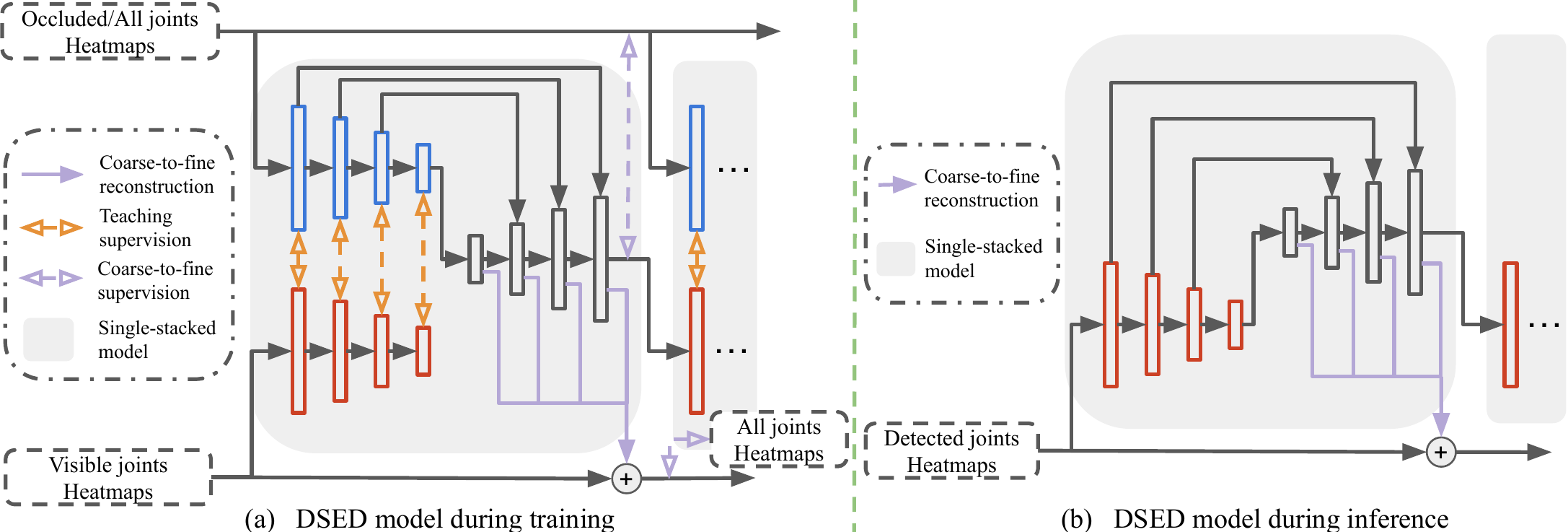}  
    \caption{\textbf{Overview of DSED network}: We only illustrate a single-stacked network here. We also use feature aggregation~\cite{li2019rethinking} but omit it for clear visualization.}
    \label{fig:GCH}
\end{figure}
% ———————————————————————————— 

The goal of this module is to infer occluded joints from visible joints. 
A straightforward idea is to reuse the stacked hourglass model~\cite{newell2016stacked,li2019rethinking}. However, our experiments show that such a model cannot solve this problem. This indicates the difficulty of identifying useful information for occlusion reconstruction from all input information. Therefore, inspired by knowledge transfer~\cite{passalis2018learning}, we propose DSED network to alleviate the difficulty of learning by splitting the task into two subtasks, reconstruct and imitate. This is done by adding an encoder to serve as a teacher to select the most useful information from occluded joints\footnote[2]{In this section, joints represent both keypoints and 3D PAFs} and supervise the second encoder to extract the same information from just visible cues.
This model is summarized in Fig.~\ref{fig:GCH}.

% ———————————————————————————— 
During training, the first encoder (blue) takes heatmaps of occluded joints as input, followed by a decoder to reconstruct these heatmaps, aiming to learn the best feature-level information needed for reconstruction. The second encoder (red) uses detected/visible joints as input and learns to extract the same information used for occlusion reconstruction, but only from visible cues. 
During inference, only the second encoder and the decoder are used. The reconstructed heatmaps of occluded joints and the detected heatmaps of visible joints are added together as the output of this module. Our experiments also show that using all joints as the input of the teacher can also improve performance, by enabling students to denoise and refine the detected joints while inferring occluded joints.

To train this model, we deeply supervise each layer of the encoders and the decoder and seek the parameters that minimize the loss function
\begin{align}
    \mathcal{L}_{reason} = \mathcal{L}_{all}+\omega_{extract}\cdot\mathcal{L}_{extract}
\end{align}
Each term of $\mathcal{L}_{all} = \lambda_k^{all}\cdot\mathcal{L}_{k}^{all}+\lambda_p^{all}\cdot\mathcal{L}_{p}^{all}$ is similar to Eq.~\ref{eq1}, but the supervision of all joints ($\hat{H}^{all}_{k}$ and $\hat{H}^{all}_p$) are provided. $\mathcal{L}_{extract}$ denotes the MSE loss for the output of each layer between these two encoders.

% ———————————————————————————— 
For root depth reasoning, we find that using the symmetry of the human torso can already yield good results. Therefore, instead of neural networks, we use a tree search method: if the pelvis is detected with high confidence, we directly use its estimate as root depth, otherwise, we search for the symmetry torso joint pair that has high-confidence estimates, and then compute the root depth based on the skeleton structure. (Please see Sup. Mat. for details.)

% ———————————————————————————— 

\noindent\textbf{Joint training with real and synthetic data.} One problem of HPE is the lack of generalization ability. Meanwhile, 3D dataset generation and multi-person motion capture under strong occlusions are expensive and challenging, which worsen this problem. To better capture the complexity of occlusions and generalize to in-the-wild data, the reasoning module is designed to be trained without image input, thus is free from the domain gap between real data and synthetic. This enables us to build a synthetic dataset with a large range of pose and occlusion distribution and train our model on it. We design two training modes to train the reasoning module alternately: the first mode uses the output of the detection module as input while the second one uses artificially generated heatmaps of visible joints from the synthetic data as input. The former makes the occlusion reasoning more targeted at what the keypoint detection module can detect and the latter ensures the generalization ability of the reasoning module. Note that Gaussian distribution is used to model uncertainties when generating supervisions $\hat{H}_k$, but the output of the detection module cannot be a perfect Gaussian distribution. When training with the first mode, the reasoning module is more likely to prune the given heatmaps of detected joints instead of predicting occluded joints. Thus we provided extra supervision $\mathcal{L}_{occ} = \lambda_k^{occ}\cdot\mathcal{L}_{k}^{occ}+\lambda_p^{occ}\cdot\mathcal{L}_{p}^{occ}$ for occluded joints to avoid converging to this trivial solution. $\mathcal{L}_{k}^{occ}$ and $\mathcal{L}_{p}^{occ}$ are similar to $\mathcal{L}_{k}^{vis}$ and $\mathcal{L}_{p}^{vis}$ in Eq.~\ref{eq1}, but only for occluded joints.

%% ===============================================
\subsection{Grouping and Refinement}
This module is not the main contribution of our method. For joint association, we directly use the depth-aware part association from SMAP~\cite{zhen2020smap} to get the connection relations, followed by a standard process to get the 3D pose from 2D joint location and relative depth. Finally a RefineNet is used to refine 3D root-relative pose. (Please refer to~\cite{zhen2020smap} for details.)

%% file: 04-occLabelGen.tex
\begin{figure}[t]
    \centering
    \includegraphics[width=0.6\columnwidth]{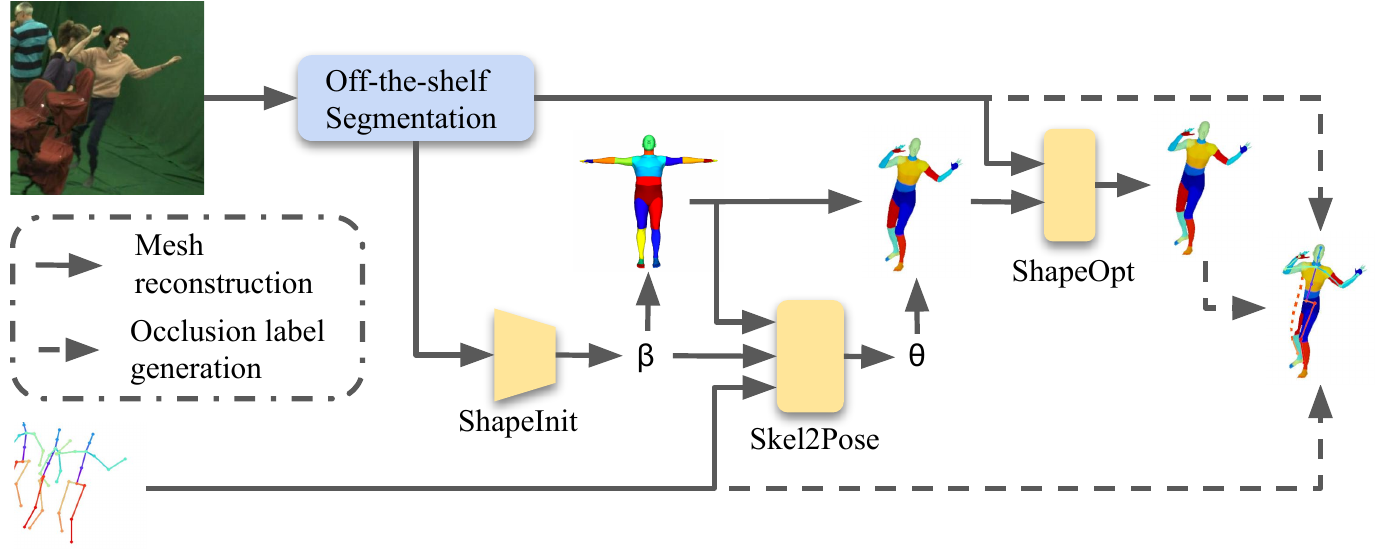}  
    \caption{\textbf{Schematic view of human shape estimation and occlusion label generation.} Test-time-optimization is used when generating occlusion labels.}
    \label{fig:shapeOpt}
\end{figure}
To explicitly learn from occluded joints, we need the occlusion label for each joint. However, existing 3D human pose datasets have no occlusion labels. Recently, Cheng~\etal~\cite{cheng2019occlusion} propose the ``Cylinder Man Model", in which they model the human body as cylindrical segments with pre-defined diameters and then project the human model into 2D to determine the degree of occlusion for each joint. This model is simple enough, but using cylinders with fixed diameters causes a very rough representation of human shape and low accuracy. More importantly, It cannot handle occlusion caused by other people and objects. Using a more complex human body model like SMPL~\cite{SMPL:2015} is a straightforward extension. However, current methods cannot provide a sufficiently accurate estimate to generate occlusion labels. To solve the problem of human shape estimation from given poses and to enable the occlusion-label generation model to accurately detect occlusion of self, objects, and other people, we proposed Skeleton-guided human Shape Fitting (SSF). Note that SSF is not our major novelty, but an essential step to generate occlusion labels (Sec.~\ref{sec:exp-ssf}) and requires extra effort.

The overall framework of this method is illustrated in Fig.~\ref{fig:shapeOpt}. Given an image $I$ and the 3D pose $P$ of all people in the image as input, an off-the-shelf instance segmentation method is used to generate instance masks $S_{ins}$ for objects and people in the image. Then for each person, we learn the shape parameters $\beta\in\mathbb{R}^{10}$ from the ShapeInit model. The shape parameters $\beta$ are then used to obtain the canonical human mesh $M_c$ with body part segmentation from the SMPL model. Next, a Skeleton2Pose model is utilized to predict the pose parameters $\theta\in\mathbb{R}^{24\times 3}$ from $M_c, \beta$, and $P$. Due to the same function and similar input, we modify the Adaptive HybrIK~\cite{li2021hybrik} and use it as the Skeleton2Pose model here. After that, we use the shape $\beta$ and the pose $\theta$ to calculate the SMPL body mesh $M_{init}=\mathcal{M}(\theta,\beta)$.  Finally, mask $S_{ins}$ and reconstructed mesh $M_{init}$ are passed through the ShapeOpt model to obtain a more accurate mesh reconstruction $M_{opt}$. More details are provided in the Sup. Mat.

The overall loss function for training this model is given by:
\begin{align}
    \mathcal{L}_{HS} = \lambda_\beta\mathcal{L}_{\beta}+\lambda_\theta\mathcal{L}_{\theta}+\lambda_{pos}\mathcal{L}_{pose}+\lambda_{sil}\mathcal{L}_{silhouette}
\end{align}
where for the $i$-th person,
\begin{align}
    \mathcal{L}_{\beta} = ||\beta^i-\hat{\beta}^i||_2 \qquad \mathcal{L}_{\theta} = ||\theta^i-\hat{\theta}^i||_2 \\
    \mathcal{L}_{pose} = ||W_{15}(\mathcal{M}(\theta^i,\beta^i))-P^i||_2^2 \\
    \mathcal{L}_{silhouette} = ||\Pi_c(\mathcal{M}(\theta^i,\beta^i))-b^i(I)||_2^2 
\end{align}
where $\hat{\beta}$ and $\hat{\theta}$ denote ground-truth shape and pose parameters respectively. $W_{15}$ is a pretrained linear regressor to output 15 joints locations consistent with the MPI15 keypoint ordering, $\Pi_c$ is the image formation function of a weak-perspective camera $c$, and $b^i(I)$ is the binary segmentation mask of person $i$ provided by the segmentation model. $\mathcal{L}_{silhouette}$ is used to reduce the error caused by the inaccurate pseudo-ground-truth SMPL annotations for in-the-wild datasets. $\mathcal{L}_{pose}$ and $\mathcal{L}_{silhouette}$ are weakly supervised loss and are also used to optimize the shape parameters when generating occlusion labels. 

After getting the human part segmentation labels $S_{part}$ of the reconstructed human mesh $M_{opt}$ from the SMPL model, we can then generate the occlusion label $o_j\in\{0,1,2\}$ of joint $j$ by checking the pixel label of $(u_j,v_j)$ given by the instance segmentation masks $S_{ins}$ and the human part segmentation labels $S_{part}$. $o_j = 0,1,2$ denote truncated, occluded, and visible, respectively.

%% file: 05-exp_v1.tex
\subsection{Dataset and Metrics}
\noindent\textbf{Dataset.}
For 3D human poes estimation (Sec.~\ref{sec:BenM-Eva}), we report results on MuPoTS-3D~\cite{mehta2018single}, CMU Panoptic~\cite{joo2015panoptic}, 3DPW~\cite{von2018recovering}, 3DPW-OCC~\cite{von2018recovering,zhang2020object}, 3DOH~\cite{zhang2020object} and Human3.6M~\cite{ionescu2013human3} datasets. Inspired by PARE~\cite{kocabas2021pare}, we randomly generate synthetic occlusions on the image of MuPoTS-3D and name it MuPoTS-synthOcc. It is only used for evaluation. For experiments on MuPoTS-3D, MuPoTS-synthOcc, 3DPW, and 3DPW-OCC, we follow SMAP~\cite{zhen2020smap} and train our model on the MuCo-3DHP~\cite{mehta2018single} dataset. In addition, for a fair comparison, we mix the data with COCO2017~\cite{lin2014microsoft} during training and $50\%$ of data in each mini-batch is from it (same as~\cite{mehta2018single,mehta2017vnect,zhen2020smap}). For Panoptic, following~\cite{zanfir2018monocular,zhen2020smap}, we choose cameras 16 and 30, and randomly select 9600 images from four activities (Haggling, Mafia, Ultimatum, Pizza) as test set, and 160k images from other sequences as training set. The synthetic dataset for training the DSED reasoning module is built based on AMASS~\cite{mahmood2019amass}. Results on Human3.6M are reported in Sup. Mat.

For the broader study of the reasoning module and the DSED network (Sec.~\ref{sec:broaderStudy}), we apply our reasoning module to two SOTA methods (PifPaf~\cite{kreiss2019pifpaf} and  HigherHRNet~\cite{cheng2020higherhrnet}) and evaluate them on COCO~\cite{lin2014microsoft} and CrowdPose~\cite{li2019crowdpose}. For a fair comparison, we directly adopt the official implementation and use the same training data. Results on CrowdPose are reported in Sup. Mat.

\noindent\textbf{Metrics.}
For pose estimation, we consider mean per joint position error (MPJPE) in mm and percentage of correct keypoints (PCK) in 3D. Following~\cite{zhen2020smap}, a keypoint is declared correct if the Euclidean distance error is smaller than 150mm. We evaluate absolute pose accuracy (subscript \textit{abs}), relative pose accuracy with root alignment (subscript \textit{rel}), and relative pose accuracy of occluded joints (subscript \textit{occ}). For human mesh reconstruction, we consider MPJPE, Procrustes-aligned mean per joint position error (PA-MPJPE), and per vertex error (PVE).

\begin{table}[t]
\renewcommand\arraystretch{0.9}
    \small 
    \centering
    \caption{\textbf{Comparisons on MuPoTS-3D and MuPoTS-synthOCC.} Results on MuPoTS-synthOcc are generated from the official pre-trained model and code if they are released. The PCK$_{occ}$ evaluates the same joints in both datasets to better analyze the occlusion reasoning ability under different levels of occlusion. Best in \textbf{bold}, second best \underline{underlined}.}
    \label{table:MuPoTS-3D}
    \resizebox{\textwidth}{!}{
    \begin{tabular}{ccccccc|ccccc}
    \toprule
    &&\multicolumn{5}{c}{MuPoTS-3D}& \multicolumn{5}{|c}{MuPoTS-synthOCC}\\
    \cmidrule(lr){3-7} \cmidrule(l){8-12}
    &&\multicolumn{3}{c}{Matched people}& \multicolumn{2}{|c}{All people} & \multicolumn{3}{|c}{Matched people}& \multicolumn{2}{|c}{All people} \\
    \cmidrule(lr){3-5} \cmidrule(lr){6-7} \cmidrule(l){8-10} \cmidrule(l){11-12} 
    & &PCK$_{abs}$&PCK$_{rel}$&PCK$_{occ}$& \multicolumn{1}{|c}{PCK$_{abs}$}&PCK$_{rel}$ &PCK$_{abs}$&PCK$_{rel}$&PCK$_{occ}$& \multicolumn{1}{|c}{PCK$_{abs}$}&PCK$_{rel}$\\
    \midrule
    \multirow{4}{*}{\shortstack{top\\down}}
    &Lcr-net++~\cite{rogez2019lcr} &-&74.0&-&-&-&-&-&-&-&-\\
    &Moon~\etal~\cite{moon2019camera} & 31.8 & 82.5 & 66.8 & 31.5 & \textbf{81.8} & 26.9 & 74.2 & 57.9 & 15.4 & 45.8\\
    &HMOR~\cite{wang2020hmor} & \textbf{43.8} & 82.0 & - & - & - & - & - & - & -  & - \\
    &HDNet~\cite{lin2020hdnet} & 35.2 & 83.7 & - & - & - & 25.8 & 72.3 & 55.9 & - & - \\
    \midrule
    \multirow{5}{*}{\shortstack{bottom\\up}}
    & ORPM~\cite{mehta2018single} & - & 69.8  & - & - & - & - & - & - & -&-\\
    & XNect~\cite{mehta2019xnect} & - & 75.8 & 57.8 & - & - & - & 69.2  & 56.2 & - & - \\
    & SMAP~\cite{zhen2020smap} & 38.7 & 80.5 & 72.9 & 35.4 & 73.5 & \underline{36.4} & 76.1 & 68.9 & 23.9 & 49.1 \\
    %&SMAP (Our trainin\underline{g)\t}extbf{!!}& 37.2 & 81.0 & 100.65 & 79.6 & 104.56 & 69.9 & 132.55 \\
    &Ours & 38.9 & \underline{84.3} & \underline{74.1} & \underline{35.8} & 76.9 & 36.3 & \underline{80.1} &  \underline{71.0} & \underline{24.3} & \underline{52.6} \\
    &Ours (w/ synthetic) & \underline{39.3} & \textbf{86.5} & \textbf{74.9} & \textbf{36.5} & \underline{79.4} & \textbf{37.9} & \textbf{81.7}  & \textbf{72.1} & \textbf{25.5} & \textbf{53.7} \\
    \bottomrule
    \end{tabular}}
\end{table}

\subsection{Benchmark Evaluation}
\label{sec:BenM-Eva}
\noindent\textbf{MuPoTS-3D and MuPoTS-synthOCC.}
Table \ref{table:MuPoTS-3D} compares our method with previous monocular HPE methods. In terms of the relative pose accuracy on which our work mainly focuses, our method significantly surpasses previous bottom-up methods in both visible and occluded joints estimate. Remarkably, even though we use a very similar detection module and the same grouping method, our method outperforms SMAP, which is also the previous SOTA bottom-up method, by 6 PCK$_{rel}$ and 5.6 PCK$_{rel}$ on matched people of MuPoTS-3D and MuPoTS-synthOCC, respectively. When considering all people, we still surpass previous art by 5.9 PCK$_{rel}$ and 4.6 PCK$_{rel}$ on these two datasets. 

More importantly, although top-down methods have the inherent advantage on accuracy since they can use off-the-shelf human detection method and can simplify the problem to single-person pose estimation, our method still outperforms them by 2.8 PCK$_{rel}$ on MuPoTS-3D when considering matched people, and is comparable to them when considering all people. Meanwhile, depending on the number of people in the image, we are faster than the SOTA top-down methods during inference. Overall, the strong results and the relatively fast inference speed prove the effectiveness and efficiency of our method.

\begin{table}[t]
    \renewcommand\arraystretch{0.9}
    \caption{\textbf{Comparisons on 3DPW, 3DPW-OCC, 3DOH, and Panoptic.} Our method yields clear improvements among all datasets. MPJPE is used.}
    \label{tab:other}
    \begin{subtable}[t]{0.43\textwidth}
        \resizebox{0.95\textwidth}{!}{
        \begin{tabular}{cccc}
        \toprule
        & 3DPW & 3DPW-OCC & 3DOH \\
        \midrule
        Moon~\etal~\cite{moon2019camera} & 98.4 & 104.3 & 89.5 \\
        XNect~\cite{mehta2019xnect} & 118.5 & 124.7 & - \\
        SMAP~\cite{zhen2020smap} & 101.5 & 105.2 & 90.6\\
        Ours & \underline{95.8} & \underline{96.9} &-\\
        Ours (w/ Synth) & \textbf{93.1} & \textbf{94.4} & \textbf{80.9} \\
        \bottomrule
        \end{tabular}}
        \caption{3DPW, 3DPW-OCC, and 3DOH}
    \end{subtable}
    \begin{subtable}[t]{.52\textwidth}
        \resizebox{0.95\textwidth}{!}{
        \begin{tabular}{cccccc}
        \toprule
         & Haggling & Mafia & Ultim. & Pizza & Average\\
        \midrule
        Moon~\etal~\cite{moon2019camera} & 89.6 & 91.3 & 79.6 & 90.1 & 87.6\\
        Zanfir~\etal~\cite{zanfir2018deep} & 72.4 & 78.8& 66.8 & 94.3 & 72.1\\
        SMAP~\cite{zhen2020smap} & 63.1 & 60.3 & 56.6 & 67.1 & 61.8 \\
        Ours &  \textbf{54.7} & \underline{55.2} & \textbf{50.1} & \underline{66.4} & \underline{56.1}  \\
        Ours (w/ Synth) & \underline{55.2} & \textbf{55.0} & \underline{50.4} & \textbf{61.4} & \textbf{55.0}\\
        \bottomrule
        \end{tabular}}
        \caption{CMU Panoptic}
    \end{subtable}
\end{table}

\noindent\textbf{3DPW, 3DPW-OCC, and 3DOH.}
These dataset are designed for multi-person human shape reconstruction. It is unfair to compare the errors between the skeleton-based method with SMPL model-based method due to the different definitions of joints. Table~\ref{tab:other} mainly focuses on skeleton-based methods. Results of baselines are generated from the official model. We use the same scripts to match the predicted person with ground-truth and compute the error based on MPI15 joint definition. Therefore, the relative values are more meaningful.

\subsection{Ablation Study}

\begin{table}[t]
\renewcommand\arraystretch{0.9}
    \small 
    \centering
    \caption{\textbf{Ablation studies on MuPoTS-3D dataset.} Det, Reason, and Ref stand for detection, reasoning, and refinement module, respectively. Hg stands for hourglass model, NL for nonlocal blocks. Det of SMAP uses all joints during training and Det of ours uses visible joints only.}
    \label{table:Abla}
    \resizebox{0.7\columnwidth}{!}{
    \begin{tabular}{lcccc}
    \toprule
    &PCK$_{rel}\uparrow$&\multicolumn{1}{c}{MPJPE$_{rel}\downarrow$ } &PCK$_{occ}\uparrow$&MPJPE$_{occ}\downarrow$\\
    \midrule
    \multicolumn{5}{c}{\textit{{Effect of each module of SMAP\cite{zhen2020smap}}}}\\
    \multicolumn{5}{c}{\tabdashline}\nextRow
    SMAP Det & 70.9 & 122.1 & 56.8 & 158.4 \\
    SMAP Det + Ref & 80.5 & 103.3 & 72.9 & 122.8 \\
    \midrule
    \multicolumn{5}{c}{\textit{{Effect of each module}}}\\
    \multicolumn{5}{c}{\tabdashline}\nextRow
    Det & 74.87 & 116.36 & 51.48 & 184.31 \\
    Det + Reason & 79.28 & 104.33 & 61.24 & 145.94\\
    Det + Reason + Ref & 86.54 & 87.28 & 74.92 & 118.60\\
    \midrule
    %\hline
    %Det (single root)\\
    %Det (multi-root)\\
    %\hline
    \multicolumn{5}{c}{\textit{{Training without occlusion label (OccL)}}}\\
    \multicolumn{5}{c}{\tabdashline}\nextRow
    Det (w/o OccL) & 71.12 & 123.02 & 56.53 & 163.74\\
    Det (w/o OccL) + Reason & 74.76 & 112.77 & 58.70 & 118.61\\
    \midrule
    \multicolumn{5}{c}{\textit{{Deeply Supervised Encoder Distillation}}}\\
    \multicolumn{5}{c}{\tabdashline}\nextRow
    Det + Reason (Hg) & 75.58 & 113.35 & 55.30 & 154.76\\
    Det + Reason (DSED) & 79.28 & 104.33 & 61.24 & 145.94\\
    Det + Reason (DSED + NL) & 79.81 & 102.32 & 61.56 & 143.59\\
    \bottomrule
    \end{tabular}}

\end{table}

\noindent\textbf{Effect of each module.}
Compared with other bottom-up methods, we add the occlusion keypoint reasoning module. First, we validate the efficiency of this module and the continuous improvement of different modules. The results can be found in \textit{Effect of each module} in Table~\ref{table:Abla}.  We achieve a continuous improvement in the accuracy of both visible and occluded joints with different modules. In addition, compared with SMAP that predicts all keypoints at the same time and then uses single-person refinement to complete the missing prediction, the idea of splitting the detection step into detection and reasoning already yields an improvement of 3.97 PCK$_{rel}$ after detection module and 8.38 PCK$_{rel}$ after reasoning module. Note that SMAP and our method use the same three-stacked hourglass architecture and output similar intermediate results.

\noindent\textbf{Training with occlusion label.} 
Next, we evaluate the effect of using occlusion labels. We can find that without using occlusion labels, the network will try to estimate both visible and occluded joints at the same time. It achieves a higher accuracy (56.53 PCK$_{occ}$ \vs 51.48 PCK$_{occ}$) on occluded joint detection but gets a lower accuracy (71.12 PCK$_{rel}$ \vs 74.87 PCK$_{rel}$) on all joints.  Training the detection module with occluded joints can to some degree improve the ability of occluded joint detection, but increasing the training difficulty and adding noises by many unpredictable joints. When the reasoning module is used in this case, the accuracy is enhanced but still lower than the model trained with occlusion labels by 4.52 PCK$_{rel}$ due to the false-positive prediction and the noise on visible keypoint prediction that is hard to be fixed by the following reasoning module.

\begin{figure}
    \centering
    \includegraphics[width=0.59\columnwidth]{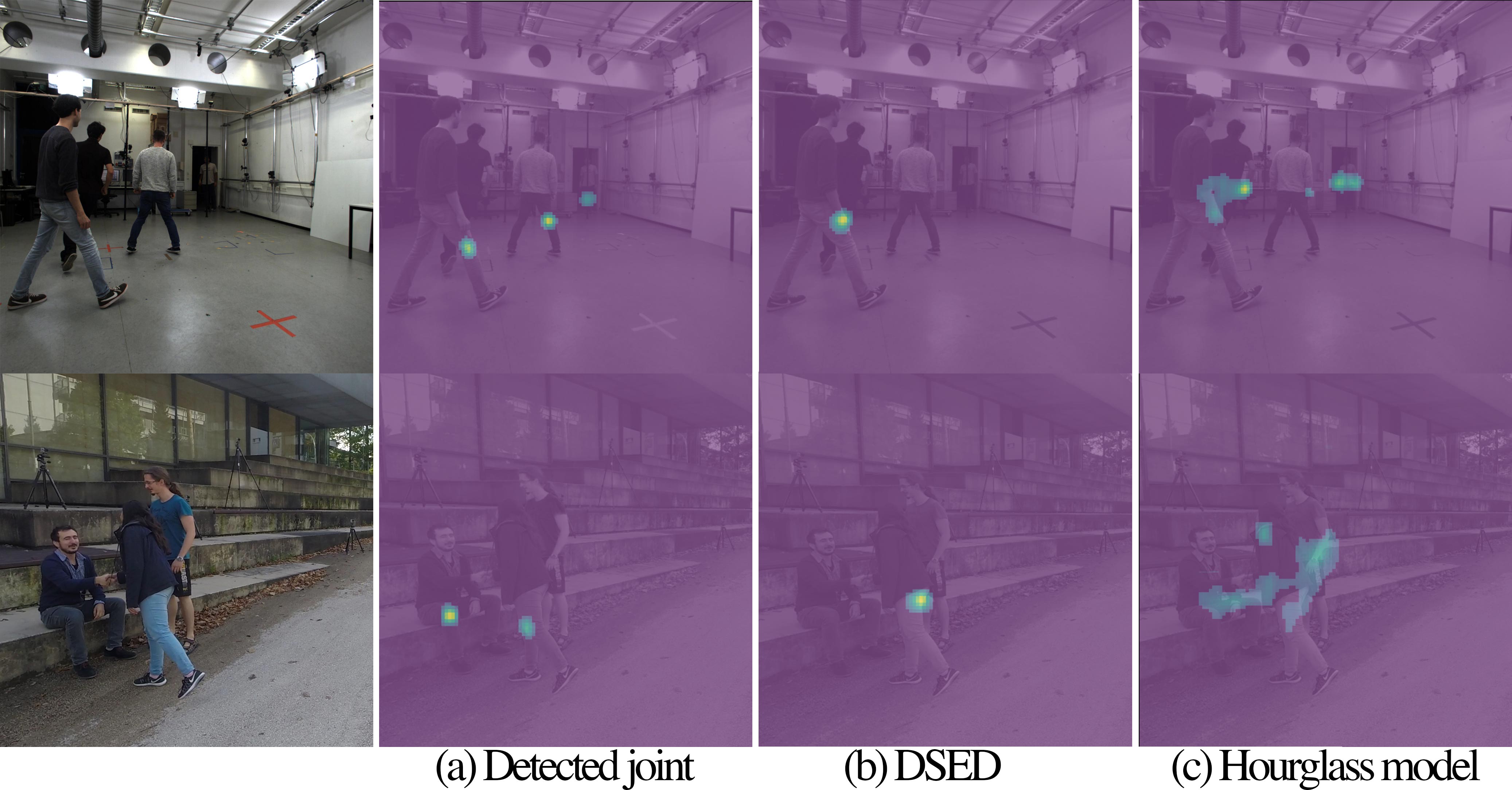}  
    \caption{\textbf{Keypoint detection and reasoning.} We visualize the right knee here. More results are provided in Sup. Mat. From left to right: input image, (a) detected keypoints, keypoints inferred by (b) DSED, and by (c) hourglass model.}
    \label{fig:KP}
\end{figure}

\noindent\textbf{DSED network.}
Compared with only using the detection model, both the hourglass-based reasoning model and DSED-based reasoning model yield improvement on occluded joints prediction. However, the DSED model achieves a much higher accuracy on occlusion reasoning by 5.94 PCK$_{occ}$. More importantly, the overall accuracy only improves 0.71 PCK$_{rel}$ when using the hourglass model but 4.41 PCK$_{rel}$ with the DSED model. This is mainly due to the false-positive estimates and noises given by the hourglass model (see Fig.~\ref{fig:KP}). 
In addition, using non-local (NL) blocks can also yield an improvement of $0.53$ PCK$_{rel}$ after carefully design, but it also increases the memory and time required for training. We do not add it to our final model but only show the results here.

\begin{table}
    \small 
    \centering
    \caption{\textbf{Broader study on 2D human pose.} We show results in single-scale testing setting. Reason stands for reasoning module, Hg for hourglass model.}
    \label{table:broader}
    \begin{subtable}[t]{0.495\textwidth}
        \resizebox{\textwidth}{!}{
        \begin{tabular}{lcccccccccc}
        \toprule
        \textit{COCO val-dev} & $\text{AP}$ & $\text{AP}^{50}$ & $\text{AP}^{75}$ & $\text{AP}^{M}$& $\text{AP}^{L}$ & $\text{AR}$  \\
        \midrule
        PifPaf~\cite{kreiss2019pifpaf} & 67.4 & - & - & - & - & - & \\
        \quad + Reason (Hg) & 67.5 & 86.5 & 73.6 & 62.0 & 75.8 & 70.9\\
        \quad + Reason (DSED) & 69.1 & 87.0 & 75.3 & 64.7 & 76.9 & 75.5\\
        \midrule
        HrHRNet-W48~\cite{cheng2020higherhrnet} & 69.9 & 87.2 & 76.1 & - & - & -  \\
        \quad + Reason (Hg) & 68.2 & 86.7 & 75.9 & 64.3 & 76.6 & 72.1\\
        \quad + Reason (DSED) & 70.8 & 87.9 & 77.0 & 66.0 & 78.3 & 76.6\\
        \bottomrule
        \end{tabular}}
        \caption{COCO val-dev 2017}
    \end{subtable}
    \begin{subtable}[t]{.495\textwidth}
        \resizebox{\textwidth}{!}{
        \begin{tabular}{lcccccccccc}
        \toprule
        \textit{COCO test-dev} & $\text{AP}$ & $\text{AP}^{50}$ & $\text{AP}^{75}$ & $\text{AP}^{M}$& $\text{AP}^{L}$ & $\text{AR}$  \\
        \midrule
        PifPaf~\cite{kreiss2019pifpaf} & 66.7 & - & - & 62.4 & 72.9 & - \\
        \quad + Reason (Hg) & 66.9 & 88.1 & 72.9 & 62.3 & 73.1 & 70.4\\
        \quad + Reason (DSED) & 68.0 & 88.7 & 75.2 & 64.1 & 74.7 & 75.6\\
        \midrule
        HrHRNet-W48~\cite{cheng2020higherhrnet} & 68.4 & 88.2 & 75.1 & 64.4 & 74.2 & -  \\
        \quad + Reason (Hg) & 67.2 &  87.4 & 74.6 & 63.0 & 73.8 & 71.8\\
        \quad + Reason (DSED) & 69.5 & 89.0 & 76.6 & 65.2 & 76.2 & 75.7\\
        \bottomrule
        \end{tabular}}
        \caption{COCO test-dev 2017}
    \end{subtable}
\end{table}

\subsection{Broader Study of the Reasoning Module and DSED Network}
\label{sec:broaderStudy}
The previous section discusses the effect of the reasoning module and DSED network in 3D HPE, now we consider a more general task, \ie, 2D human pose estimation. We delete the depth-related structure of the reasoning module and modify the DSED network, then apply them to PifPaf and HigherHRNet. We use the official implementation with small modifications on their network to make it compatible with our 2D reasoning module (see Sup. Mat. for details). The results are reported in Table~\ref{table:broader}. All methods are trained on 2017 COCO training set. We can see that even though the reasoning module is not designed for 2D tasks since the bone length is more accurate when computed in 3D, it still yields stable improvement (up to 1.7 AP) and achieves SOTA performances. We can also find that DSED is crucial for the reasoning module. 

\begin{table}
    \setlength{\abovecaptionskip}{0.05cm}
    \setlength{\belowcaptionskip}{-0.15cm}
    \renewcommand\arraystretch{0.5}
    \renewcommand{\baselinestretch}{0.8}
    \small 
    \centering
    \caption{\textbf{Evaluation of occluded joint detection.} The left half compares the generated occlusion labels and the right half shows the model performance trained on these labels. For the left half, we randomly select 500 from 200k images in MuCo-3DHP and manually annotate them to get ground-truth labels.}
    \label{table:occL}
    \resizebox{0.7\columnwidth}{!}{
    \begin{tabular}{cccc|ccc}
    \toprule
     & Precision & Recall & F1-score & \multicolumn{1}{|c}{PCK$_{abs}$}& PCK$_{rel}$ &PCK$_{occ}$\\
    \midrule
    Cylinder~\cite{cheng2019occlusion} & 0.653 & 0.401 & 0.497 & - & - & -  \\
    HybrIK-based~\cite{li2021hybrik} & 0.800 & 0.602 & 0.687 & 38.4 & 83.1 & 73.0  \\
    ours (SSF-based) & 0.885 & 0.805 & 0.843 & \textbf{39.3} & \textbf{86.5} & \textbf{74.9} \\
    \bottomrule
    \end{tabular}}
\end{table}

\subsection{SSF and Occlusion Label Generation}
\label{sec:exp-ssf}
\noindent\textbf{Occlusion label generation.}
We compare our method with the Cylinder human model and the SMPL-based human mesh fitting method (\ie, Adaptive HybrIK). We implement the Cylinder model and modify it to detect occlusion by other people. For comparison with HybrIK, we directly use the official model to fit the SMPL models and then use the same graphics pipeline in our SSF to generate the occlusion labels. Note that the ground-truth skeleton is provided for all methods. The results can be found in Table~\ref{table:occL}. Our method generates 
more accurate occlusion labels with much higher precision and recall. More importantly, the labels generated by previous methods reduce performance. SSF is essential to generate occlusion labels for explicit occlusion reasoning.

\begin{table}
\renewcommand\arraystretch{0.9}
    \setlength{\abovecaptionskip}{0.05cm}
    \setlength{\belowcaptionskip}{-0.15cm}
    \renewcommand\arraystretch{0.5}
    \renewcommand{\baselinestretch}{0.8}
    \small
    \centering
    \caption{\textbf{Evaluation of human mesh recovery on 3DPW dataset.} }
    \label{table:HSE}
    \resizebox{0.6\columnwidth}{!}{
    \begin{tabular}{cccc}
    \toprule
    & MPJPE & PA-MPJPE & PVE\\
    \midrule
    SPIN~\cite{kolotouros2019learning} & 96.9 & 59.2 & 116.4\\
    ROMP (ResNet-50)~\cite{sun2021monocular} & 91.3 & 54.9 & 108.3 \\
    PARE (ResNet-50)~\cite{kocabas2021pare} & 84.3 & 51.2 & 101.2\\
    Adaptive HybrIK (ResNet-34)~\cite{li2021hybrik} & \underline{80.0} & \textbf{48.8} & \underline{94.5} \\
    Ours (w/o Synth, ResNet-34) & \textbf{79.1} & \underline{49.3} & \textbf{92.3}\\
    \bottomrule
    \end{tabular}}
\end{table}

\noindent\textbf{3D human mesh reconstruction.}
We evaluate our SSF on the 3DPW dataset. We use the proposed HUPOR to generate 3D skeletons, and then use the estimated skeletons and the image as input to estimate SMPL parameters. For a fair comparison, we strictly follow HybrIK~\cite{li2021hybrik} to prepare the training data and use ResNet-34~\cite{he2016deep} as backbone. The results are reported in Table~\ref{table:HSE}. Compared with baselines, SSF can recover body mesh more accurately. Qualitative results on human shape estimation can be found in Sup. Mat.

%% file: eccv2022submission.bbl
\begin{thebibliography}{10}
\providecommand{\url}[1]{\texttt{#1}}
\providecommand{\urlprefix}{URL }
\providecommand{\doi}[1]{https://doi.org/#1}

\bibitem{akhter2015pose}
Akhter, I., Black, M.J.: Pose-conditioned joint angle limits for 3d human pose
  reconstruction. In: Proceedings of the IEEE conference on computer vision and
  pattern recognition. pp. 1446--1455 (2015)

\bibitem{artacho2020unipose}
Artacho, B., Savakis, A.: Unipose: Unified human pose estimation in single
  images and videos. In: Proceedings of the IEEE/CVF Conference on Computer
  Vision and Pattern Recognition. pp. 7035--7044 (2020)

\bibitem{de2018deep}
de~Bem, R., Arnab, A., Golodetz, S., Sapienza, M., Torr, P.: Deep
  fully-connected part-based models for human pose estimation. In: Asian
  Conference on Machine Learning. pp. 327--342. PMLR (2018)

\bibitem{biggs20203d}
Biggs, B., Ehrhadt, S., Joo, H., Graham, B., Vedaldi, A., Novotny, D.: 3d
  multi-bodies: Fitting sets of plausible 3d human models to ambiguous image
  data. Advances in Neural Information Processing Systems  (2020)

\bibitem{braso2021center}
Bras{\'o}, G., Kister, N., Leal-Taix{\'e}, L.: The center of attention:
  Center-keypoint grouping via attention for multi-person pose estimation. In:
  Proceedings of the IEEE/CVF International Conference on Computer Vision. pp.
  11853--11863 (2021)

\bibitem{cai2020learning}
Cai, Y., Wang, Z., Luo, Z., Yin, B., Du, A., Wang, H., Zhang, X., Zhou, X.,
  Zhou, E., Sun, J.: Learning delicate local representations for multi-person
  pose estimation. In: European Conference on Computer Vision. pp. 455--472.
  Springer (2020)

\bibitem{cai2019exploiting}
Cai, Y., Ge, L., Liu, J., Cai, J., Cham, T.J., Yuan, J., Thalmann, N.M.:
  Exploiting spatial-temporal relationships for 3d pose estimation via graph
  convolutional networks. In: Proceedings of the IEEE/CVF International
  Conference on Computer Vision. pp. 2272--2281 (2019)

\bibitem{cao2017realtime}
Cao, Z., Simon, T., Wei, S.E., Sheikh, Y.: Realtime multi-person 2d pose
  estimation using part affinity fields. In: Proceedings of the IEEE conference
  on computer vision and pattern recognition. pp. 7291--7299 (2017)

\bibitem{chen20173d}
Chen, C.H., Ramanan, D.: 3d human pose estimation= 2d pose estimation+
  matching. In: Proceedings of the IEEE Conference on Computer Vision and
  Pattern Recognition. pp. 7035--7043 (2017)

\bibitem{chen2018cascaded}
Chen, Y., Wang, Z., Peng, Y., Zhang, Z., Yu, G., Sun, J.: Cascaded pyramid
  network for multi-person pose estimation. In: Proceedings of the IEEE
  conference on computer vision and pattern recognition. pp. 7103--7112 (2018)

\bibitem{cheng2020higherhrnet}
Cheng, B., Xiao, B., Wang, J., Shi, H., Huang, T.S., Zhang, L.: Higherhrnet:
  Scale-aware representation learning for bottom-up human pose estimation. In:
  Proceedings of the IEEE/CVF Conference on Computer Vision and Pattern
  Recognition. pp. 5386--5395 (2020)

\bibitem{cheng20203d}
Cheng, Y., Yang, B., Wang, B., Tan, R.T.: 3d human pose estimation using
  spatio-temporal networks with explicit occlusion training. In: Proceedings of
  the AAAI Conference on Artificial Intelligence. vol.~34, pp. 10631--10638
  (2020)

\bibitem{cheng2019occlusion}
Cheng, Y., Yang, B., Wang, B., Yan, W., Tan, R.T.: Occlusion-aware networks for
  3d human pose estimation in video. In: Proceedings of the IEEE/CVF
  International Conference on Computer Vision. pp. 723--732 (2019)

\bibitem{choi2020pose2mesh}
Choi, H., Moon, G., Lee, K.M.: Pose2mesh: Graph convolutional network for 3d
  human pose and mesh recovery from a 2d human pose. In: European Conference on
  Computer Vision. pp. 769--787. Springer (2020)

\bibitem{chu2017multi}
Chu, X., Yang, W., Ouyang, W., Ma, C., Yuille, A.L., Wang, X.: Multi-context
  attention for human pose estimation. In: Proceedings of the IEEE conference
  on computer vision and pattern recognition. pp. 1831--1840 (2017)

\bibitem{fang2017rmpe}
Fang, H.S., Xie, S., Tai, Y.W., Lu, C.: Rmpe: Regional multi-person pose
  estimation. In: Proceedings of the IEEE international conference on computer
  vision. pp. 2334--2343 (2017)

\bibitem{gu2021exploring}
Gu, R., Wang, G., Hwang, J.N.: Exploring severe occlusion: Multi-person 3d pose
  estimation with gated convolution. In: 2020 25th International Conference on
  Pattern Recognition (ICPR). pp. 8243--8250. IEEE (2021)

\bibitem{he2017mask}
He, K., Gkioxari, G., Doll{\'a}r, P., Girshick, R.: Mask r-cnn. In: Proceedings
  of the IEEE international conference on computer vision. pp. 2961--2969
  (2017)

\bibitem{he2016deep}
He, K., Zhang, X., Ren, S., Sun, J.: Deep residual learning for image
  recognition. In: Proceedings of the IEEE conference on computer vision and
  pattern recognition. pp. 770--778 (2016)

\bibitem{huang2020devil}
Huang, J., Zhu, Z., Guo, F., Huang, G.: The devil is in the details: Delving
  into unbiased data processing for human pose estimation. In: Proceedings of
  the IEEE/CVF Conference on Computer Vision and Pattern Recognition. pp.
  5700--5709 (2020)

\bibitem{huang2017coarse}
Huang, S., Gong, M., Tao, D.: A coarse-fine network for keypoint localization.
  In: Proceedings of the IEEE international conference on computer vision. pp.
  3028--3037 (2017)

\bibitem{insafutdinov2016deepercut}
Insafutdinov, E., Pishchulin, L., Andres, B., Andriluka, M., Schiele, B.:
  Deepercut: A deeper, stronger, and faster multi-person pose estimation model.
  In: European Conference on Computer Vision. pp. 34--50. Springer (2016)

\bibitem{ionescu2013human3}
Ionescu, C., Papava, D., Olaru, V., Sminchisescu, C.: Human3. 6m: Large scale
  datasets and predictive methods for 3d human sensing in natural environments.
  IEEE transactions on pattern analysis and machine intelligence
  \textbf{36}(7),  1325--1339 (2013)

\bibitem{jahangiri2017generating}
Jahangiri, E., Yuille, A.L.: Generating multiple diverse hypotheses for human
  3d pose consistent with 2d joint detections. In: Proceedings of the IEEE
  International Conference on Computer Vision Workshops. pp. 805--814 (2017)

\bibitem{jin2020differentiable}
Jin, S., Liu, W., Xie, E., Wang, W., Qian, C., Ouyang, W., Luo, P.:
  Differentiable hierarchical graph grouping for multi-person pose estimation.
  In: European Conference on Computer Vision. pp. 718--734. Springer (2020)

\bibitem{joo2015panoptic}
Joo, H., Liu, H., Tan, L., Gui, L., Nabbe, B., Matthews, I., Kanade, T.,
  Nobuhara, S., Sheikh, Y.: Panoptic studio: A massively multiview system for
  social motion capture. In: Proceedings of the IEEE International Conference
  on Computer Vision. pp. 3334--3342 (2015)

\bibitem{joo2020exemplar}
Joo, H., Neverova, N., Vedaldi, A.: Exemplar fine-tuning for 3d human model
  fitting towards in-the-wild 3d human pose estimation. arXiv preprint
  arXiv:2004.03686  (2020)

\bibitem{kanazawa2018end}
Kanazawa, A., Black, M.J., Jacobs, D.W., Malik, J.: End-to-end recovery of
  human shape and pose. In: Proceedings of the IEEE conference on computer
  vision and pattern recognition. pp. 7122--7131 (2018)

\bibitem{khirodkar2021multi}
Khirodkar, R., Chari, V., Agrawal, A., Tyagi, A.: Multi-hypothesis pose
  networks: Rethinking top-down pose estimation. Proceedings of the IEEE
  international conference on computer vision pp. 3122--3131 (2021)

\bibitem{kocabas2021pare}
Kocabas, M., Huang, C.H.P., Hilliges, O., Black, M.J.: Pare: Part attention
  regressor for 3d human body estimation. Proceedings of the IEEE International
  Conference on Computer Vision  (2021)

\bibitem{kolotouros2019learning}
Kolotouros, N., Pavlakos, G., Black, M.J., Daniilidis, K.: Learning to
  reconstruct 3d human pose and shape via model-fitting in the loop. In:
  Proceedings of the IEEE/CVF International Conference on Computer Vision. pp.
  2252--2261 (2019)

\bibitem{kreiss2019pifpaf}
Kreiss, S., Bertoni, L., Alahi, A.: Pifpaf: Composite fields for human pose
  estimation. In: Proceedings of the IEEE/CVF Conference on Computer Vision and
  Pattern Recognition. pp. 11977--11986 (2019)

\bibitem{li2019generating}
Li, C., Lee, G.H.: Generating multiple hypotheses for 3d human pose estimation
  with mixture density network. In: Proceedings of the IEEE/CVF Conference on
  Computer Vision and Pattern Recognition. pp. 9887--9895 (2019)

\bibitem{li2020simple}
Li, J., Su, W., Wang, Z.: Simple pose: Rethinking and improving a bottom-up
  approach for multi-person pose estimation. In: Proceedings of the AAAI
  conference on artificial intelligence. vol.~34, pp. 11354--11361 (2020)

\bibitem{li2019crowdpose}
Li, J., Wang, C., Zhu, H., Mao, Y., Fang, H.S., Lu, C.: Crowdpose: Efficient
  crowded scenes pose estimation and a new benchmark. In: Proceedings of the
  IEEE/CVF Conference on Computer Vision and Pattern Recognition. pp.
  10863--10872 (2019)

\bibitem{li2021hybrik}
Li, J., Xu, C., Chen, Z., Bian, S., Yang, L., Lu, C.: Hybrik: A hybrid
  analytical-neural inverse kinematics solution for 3d human pose and shape
  estimation. In: Proceedings of the IEEE/CVF Conference on Computer Vision and
  Pattern Recognition. pp. 3383--3393 (2021)

\bibitem{li2019rethinking}
Li, W., Wang, Z., Yin, B., Peng, Q., Du, Y., Xiao, T., Yu, G., Lu, H., Wei, Y.,
  Sun, J.: Rethinking on multi-stage networks for human pose estimation. arXiv
  preprint arXiv:1901.00148  (2019)

\bibitem{lin2020hdnet}
Lin, J., Lee, G.H.: Hdnet: Human depth estimation for multi-person camera-space
  localization. In: European Conference on Computer Vision. pp. 633--648.
  Springer (2020)

\bibitem{lin2014microsoft}
Lin, T.Y., Maire, M., Belongie, S., Hays, J., Perona, P., Ramanan, D.,
  Doll{\'a}r, P., Zitnick, C.L.: Microsoft coco: Common objects in context. In:
  European conference on computer vision. pp. 740--755. Springer (2014)

\bibitem{liu2021graph}
Liu, J., Rojas, J., Li, Y., Liang, Z., Guan, Y., Xi, N., Zhu, H.: A graph
  attention spatio-temporal convolutional network for 3d human pose estimation
  in video. In: 2021 IEEE International Conference on Robotics and Automation
  (ICRA). pp. 3374--3380. IEEE (2021)

\bibitem{SMPL:2015}
Loper, M., Mahmood, N., Romero, J., Pons-Moll, G., Black, M.J.: {SMPL}: A
  skinned multi-person linear model. ACM Trans. Graphics (Proc. SIGGRAPH Asia)
  \textbf{34}(6),  248:1--248:16 (Oct 2015)

\bibitem{mahmood2019amass}
Mahmood, N., Ghorbani, N., Troje, N.F., Pons-Moll, G., Black, M.J.: Amass:
  Archive of motion capture as surface shapes. In: Proceedings of the IEEE/CVF
  International Conference on Computer Vision. pp. 5442--5451 (2019)

\bibitem{von2018recovering}
von Marcard, T., Henschel, R., Black, M.J., Rosenhahn, B., Pons-Moll, G.:
  Recovering accurate 3d human pose in the wild using imus and a moving camera.
  In: Proceedings of the European Conference on Computer Vision (ECCV). pp.
  601--617 (2018)

\bibitem{martinez2017simple}
Martinez, J., Hossain, R., Romero, J., Little, J.J.: A simple yet effective
  baseline for 3d human pose estimation. In: Proceedings of the IEEE
  International Conference on Computer Vision. pp. 2640--2649 (2017)

\bibitem{mehta2017monocular}
Mehta, D., Rhodin, H., Casas, D., Fua, P., Sotnychenko, O., Xu, W., Theobalt,
  C.: Monocular 3d human pose estimation in the wild using improved cnn
  supervision. In: 2017 international conference on 3D vision (3DV). pp.
  506--516. IEEE (2017)

\bibitem{mehta2019xnect}
Mehta, D., Sotnychenko, O., Mueller, F., Xu, W., Elgharib, M., Fua, P., Seidel,
  H.P., Rhodin, H., Pons-Moll, G., Theobalt, C.: Xnect: Real-time multi-person
  3d human pose estimation with a single rgb camera. ACM Transactions on
  Graphics (TOG)  (2020)

\bibitem{mehta2018single}
Mehta, D., Sotnychenko, O., Mueller, F., Xu, W., Sridhar, S., Pons-Moll, G.,
  Theobalt, C.: Single-shot multi-person 3d pose estimation from monocular rgb.
  In: 2018 International Conference on 3D Vision (3DV). pp. 120--130. IEEE
  (2018)

\bibitem{mehta2017vnect}
Mehta, D., Sridhar, S., Sotnychenko, O., Rhodin, H., Shafiei, M., Seidel, H.P.,
  Xu, W., Casas, D., Theobalt, C.: Vnect: Real-time 3d human pose estimation
  with a single rgb camera. ACM Transactions on Graphics (TOG)  \textbf{36}(4),
   1--14 (2017)

\bibitem{moon2019camera}
Moon, G., Chang, J.Y., Lee, K.M.: Camera distance-aware top-down approach for
  3d multi-person pose estimation from a single rgb image. In: Proceedings of
  the IEEE/CVF International Conference on Computer Vision. pp. 10133--10142
  (2019)

\bibitem{moon2019posefix}
Moon, G., Chang, J.Y., Lee, K.M.: Posefix: Model-agnostic general human pose
  refinement network. In: Proceedings of the IEEE/CVF Conference on Computer
  Vision and Pattern Recognition. pp. 7773--7781 (2019)

\bibitem{moreno20173d}
Moreno-Noguer, F.: 3d human pose estimation from a single image via distance
  matrix regression. In: Proceedings of the IEEE Conference on Computer Vision
  and Pattern Recognition. pp. 2823--2832 (2017)

\bibitem{newell2016associative}
Newell, A., Huang, Z., Deng, J.: Associative embedding: End-to-end learning for
  joint detection and grouping. arXiv preprint arXiv:1611.05424  (2016)

\bibitem{newell2016stacked}
Newell, A., Yang, K., Deng, J.: Stacked hourglass networks for human pose
  estimation. In: European conference on computer vision. pp. 483--499.
  Springer (2016)

\bibitem{papandreou2018personlab}
Papandreou, G., Zhu, T., Chen, L.C., Gidaris, S., Tompson, J., Murphy, K.:
  Personlab: Person pose estimation and instance segmentation with a bottom-up,
  part-based, geometric embedding model. In: Proceedings of the European
  Conference on Computer Vision (ECCV). pp. 269--286 (2018)

\bibitem{papandreou2017towards}
Papandreou, G., Zhu, T., Kanazawa, N., Toshev, A., Tompson, J., Bregler, C.,
  Murphy, K.: Towards accurate multi-person pose estimation in the wild. In:
  Proceedings of the IEEE conference on computer vision and pattern
  recognition. pp. 4903--4911 (2017)

\bibitem{parger2021unoc}
Parger, M., Tang, C., Xu, Y., Twigg, C.D., Tao, L., Li, Y., Wang, R.,
  Steinberger, M.: Unoc: Understanding occlusion for embodied presence in
  virtual reality. IEEE Transactions on Visualization and Computer Graphics
  (2021)

\bibitem{park2021localizing}
Park, S., Park, J.: Localizing human keypoints beyond the bounding box. In:
  Proceedings of the IEEE/CVF International Conference on Computer Vision. pp.
  1602--1611 (2021)

\bibitem{passalis2018learning}
Passalis, N., Tefas, A.: Learning deep representations with probabilistic
  knowledge transfer. In: Proceedings of the European Conference on Computer
  Vision (ECCV). pp. 268--284 (2018)

\bibitem{pavlakos2017coarse}
Pavlakos, G., Zhou, X., Derpanis, K.G., Daniilidis, K.: Coarse-to-fine
  volumetric prediction for single-image 3d human pose. In: Proceedings of the
  IEEE conference on computer vision and pattern recognition. pp. 7025--7034
  (2017)

\bibitem{pavllo20193d}
Pavllo, D., Feichtenhofer, C., Grangier, D., Auli, M.: 3d human pose estimation
  in video with temporal convolutions and semi-supervised training. In:
  Proceedings of the IEEE/CVF Conference on Computer Vision and Pattern
  Recognition. pp. 7753--7762 (2019)

\bibitem{peng2018jointly}
Peng, X., Tang, Z., Yang, F., Feris, R.S., Metaxas, D.: Jointly optimize data
  augmentation and network training: Adversarial data augmentation in human
  pose estimation. In: Proceedings of the IEEE Conference on Computer Vision
  and Pattern Recognition. pp. 2226--2234 (2018)

\bibitem{pishchulin2016deepcut}
Pishchulin, L., Insafutdinov, E., Tang, S., Andres, B., Andriluka, M., Gehler,
  P.V., Schiele, B.: Deepcut: Joint subset partition and labeling for multi
  person pose estimation. In: Proceedings of the IEEE conference on computer
  vision and pattern recognition. pp. 4929--4937 (2016)

\bibitem{qiu2020peeking}
Qiu, L., Zhang, X., Li, Y., Li, G., Wu, X., Xiong, Z., Han, X., Cui, S.:
  Peeking into occluded joints: A novel framework for crowd pose estimation.
  In: European Conference on Computer Vision. pp. 488--504. Springer (2020)

\bibitem{radwan2013monocular}
Radwan, I., Dhall, A., Goecke, R.: Monocular image 3d human pose estimation
  under self-occlusion. In: Proceedings of the IEEE International Conference on
  Computer Vision. pp. 1888--1895 (2013)

\bibitem{rogez2017lcr}
Rogez, G., Weinzaepfel, P., Schmid, C.: Lcr-net:
  Localization-classification-regression for human pose. In: Proceedings of the
  IEEE Conference on Computer Vision and Pattern Recognition. pp. 3433--3441
  (2017)

\bibitem{rogez2019lcr}
Rogez, G., Weinzaepfel, P., Schmid, C.: Lcr-net++: Multi-person 2d and 3d pose
  detection in natural images. IEEE transactions on pattern analysis and
  machine intelligence  \textbf{42}(5),  1146--1161 (2019)

\bibitem{su2019multi}
Su, K., Yu, D., Xu, Z., Geng, X., Wang, C.: Multi-person pose estimation with
  enhanced channel-wise and spatial information. In: Proceedings of the
  IEEE/CVF Conference on Computer Vision and Pattern Recognition. pp.
  5674--5682 (2019)

\bibitem{sun2019deep}
Sun, K., Xiao, B., Liu, D., Wang, J.: Deep high-resolution representation
  learning for human pose estimation. In: Proceedings of the IEEE/CVF
  Conference on Computer Vision and Pattern Recognition. pp. 5693--5703 (2019)

\bibitem{sun2017compositional}
Sun, X., Shang, J., Liang, S., Wei, Y.: Compositional human pose regression.
  In: Proceedings of the IEEE International Conference on Computer Vision. pp.
  2602--2611 (2017)

\bibitem{sun2021monocular}
Sun, Y., Bao, Q., Liu, W., Fu, Y., Black, M.J., Mei, T.: Monocular, one-stage,
  regression of multiple 3d people. In: Proceedings of the IEEE/CVF
  International Conference on Computer Vision. pp. 11179--11188 (2021)

\bibitem{veges2020temporal}
V{\'e}ges, M., L{\H{o}}rincz, A.: Temporal smoothing for 3d human pose
  estimation and localization for occluded people. In: International Conference
  on Neural Information Processing. pp. 557--568. Springer (2020)

\bibitem{wang2020hmor}
Wang, C., Li, J., Liu, W., Qian, C., Lu, C.: Hmor: Hierarchical multi-person
  ordinal relations for monocular multi-person 3d pose estimation. In: European
  Conference on Computer Vision. pp. 242--259. Springer (2020)

\bibitem{wang2020graph}
Wang, J., Long, X., Gao, Y., Ding, E., Wen, S.: Graph-pcnn: Two stage human
  pose estimation with graph pose refinement. In: European Conference on
  Computer Vision. pp. 492--508. Springer (2020)

\bibitem{wang20203d}
Wang, J., Xu, E., Xue, K., Kidzinski, L.: 3d pose detection in videos: Focusing
  on occlusion. arXiv preprint arXiv:2006.13517  (2020)

\bibitem{wehrbein2021probabilistic}
Wehrbein, T., Rudolph, M., Rosenhahn, B., Wandt, B.: Probabilistic monocular 3d
  human pose estimation with normalizing flows. In: Proceedings of the IEEE/CVF
  International Conference on Computer Vision. pp. 11199--11208 (2021)

\bibitem{xiao2018simple}
Xiao, B., Wu, H., Wei, Y.: Simple baselines for human pose estimation and
  tracking. In: Proceedings of the European conference on computer vision
  (ECCV). pp. 466--481 (2018)

\bibitem{xu2020deep}
Xu, J., Yu, Z., Ni, B., Yang, J., Yang, X., Zhang, W.: Deep kinematics analysis
  for monocular 3d human pose estimation. In: Proceedings of the IEEE/CVF
  Conference on Computer Vision and Pattern Recognition. pp. 899--908 (2020)

\bibitem{xu2021monocular}
Xu, Y., Wang, W., Liu, T., Liu, X., Xie, J., Zhu, S.C.: Monocular 3d pose
  estimation via pose grammar and data augmentation. IEEE Transactions on
  Pattern Analysis and Machine Intelligence  (2021)

\bibitem{yang20183d}
Yang, W., Ouyang, W., Wang, X., Ren, J., Li, H., Wang, X.: 3d human pose
  estimation in the wild by adversarial learning. In: Proceedings of the IEEE
  Conference on Computer Vision and Pattern Recognition. pp. 5255--5264 (2018)

\bibitem{zanfir2018monocular}
Zanfir, A., Marinoiu, E., Sminchisescu, C.: Monocular 3d pose and shape
  estimation of multiple people in natural scenes-the importance of multiple
  scene constraints. In: Proceedings of the IEEE Conference on Computer Vision
  and Pattern Recognition. pp. 2148--2157 (2018)

\bibitem{zanfir2018deep}
Zanfir, A., Marinoiu, E., Zanfir, M., Popa, A.I., Sminchisescu, C.: Deep
  network for the integrated 3d sensing of multiple people in natural images.
  Advances in Neural Information Processing Systems  \textbf{31},  8410--8419
  (2018)

\bibitem{zhang2020distribution}
Zhang, F., Zhu, X., Dai, H., Ye, M., Zhu, C.: Distribution-aware coordinate
  representation for human pose estimation. In: Proceedings of the IEEE/CVF
  conference on computer vision and pattern recognition. pp. 7093--7102 (2020)

\bibitem{zhang2020object}
Zhang, T., Huang, B., Wang, Y.: Object-occluded human shape and pose estimation
  from a single color image. In: Proceedings of the IEEE/CVF Conference on
  Computer Vision and Pattern Recognition. pp. 7376--7385 (2020)

\bibitem{zhen2020smap}
Zhen, J., Fang, Q., Sun, J., Liu, W., Jiang, W., Bao, H., Zhou, X.: Smap:
  Single-shot multi-person absolute 3d pose estimation. In: European Conference
  on Computer Vision. pp. 550--566. Springer (2020)

\bibitem{zhou2020occlusion}
Zhou, L., Chen, Y., Gao, Y., Wang, J., Lu, H.: Occlusion-aware siamese network
  for human pose estimation. In: European Conference on Computer Vision. pp.
  396--412. Springer (2020)

\bibitem{zhou2017towards}
Zhou, X., Huang, Q., Sun, X., Xue, X., Wei, Y.: Towards 3d human pose
  estimation in the wild: a weakly-supervised approach. In: Proceedings of the
  IEEE International Conference on Computer Vision. pp. 398--407 (2017)

\bibitem{zhou2019objects}
Zhou, X., Wang, D., Kr{\"a}henb{\"u}hl, P.: Objects as points. arXiv preprint
  arXiv:1904.07850  (2019)

\end{thebibliography}
